%% file: TSWLatexianTemp_main.tex
\documentclass[10pt,twocolumn,letterpaper]{article}
\usepackage[justification = centering]{caption}
\usepackage{setspace}
\usepackage{caption}
\usepackage{cvpr}
\usepackage{times}
\usepackage{epsfig}
\usepackage{graphicx}
\usepackage{amsmath}
\usepackage{microtype}
\usepackage[american]{babel}
\usepackage{amssymb}
\usepackage[numbers,sort&compress]{natbib}
\usepackage[justification=centering]{caption}

\usepackage{color}

\usepackage[pagebackref=true,breaklinks=true,letterpaper=true,colorlinks,bookmarks=false]{hyperref}

\cvprfinalcopy % *** Uncomment this line for the final submission

\def\cvprPaperID{9} % *** Enter the CVPR Paper ID here

\ifcvprfinal\fi
\begin{document}

%%%%%%%%% TITLE
\title{Tips and Tricks for Webly-Supervised Fine-Grained Recognition:\\Learning from the WebFG 2020 Challenge}

\maketitle

\begin{abstract}
WebFG 2020 is an international challenge hosted by Nanjing University of Science and Technology, University of Edinburgh, Nanjing University, The University of Adelaide, Waseda University, etc. This challenge mainly pays attention to the webly-supervised fine-grained recognition problem. In the literature, existing deep learning methods highly rely on large-scale and high-quality labeled training data, which poses a limitation to their practicability and scalability in real world applications. In particular, for fine-grained recognition, a visual task that requires professional knowledge for labeling, the cost of acquiring labeled training data is quite high. It causes extreme difficulties to obtain a large amount of high-quality training data. Therefore, utilizing free web data to train fine-grained recognition models has attracted increasing attentions from researchers in the fine-grained community. This challenge expects participants to develop webly-supervised fine-grained recognition methods, which leverages web images in training fine-grained recognition models to ease the extreme dependence of deep learning methods on large-scale manually labeled datasets and to enhance their practicability and scalability. In this technical report, we have pulled together the top WebFG 2020 solutions of total 54 competing teams, and discuss what methods worked best across the set of winning teams, and what surprisingly did not help.
\end{abstract}

\section{Introduction}

Learning from the web can ease the extreme dependence of deep learning on large-scale manually labeled datasets. Especially for fine-grained recognition~\cite{wei2019deepsurvey}, which targets at distinguishing subordinate categories, it will significantly reduce the labeling costs by leveraging free web data. However, such free data in the real-world consists of three main challenges: (1) label noises, (2) small inter-class variances and (3) extreme class imbalance. In order to explore whether this data could be leveraged to improve generalization on fine-grained image recognition, we organize a webly-supervsied fine-grained recognition challenge/workshop (WebFG 2020), which is associated with the leading biennial international conference ACCV 2020 and specifically construct a large-scale webly fine-grained dataset (5,000 sub-categories and more than 0.5 million web data) as training data to perform the test bed of the challenge.

In the WebFG 2020 workshop, participants have to deal with the challenging subordinate fine-grained category classification task (totally 5,000 sub-categories) by leveraging the free but noisy web data (a.k.a. webly fine-grained images). This results in a more robust evaluation of the current state of the art, counteracts overfitting to traditional fine-grained benchmarks, and allows insights into prevalent dataset bias, data noises and as well as long-tailed distribution of the webly fine-grained nature. The overall goal of the challenge is to evaluate the robustness and generalization capabilities of state-of-the-art algorithms. The WebFG 2020 challenge associated with ACCV 2020 has attracted 54 competing teams, hailing from all over the globe. In this technical report, we have pulled together the top-3 solutions and discuss what methods worked best across the set of wining teams, and what surprisingly did not help. We hope the technical report could benefit the future research in the line of fine-grained recognition, webly-supervised learning and long-tailed recognition.

\section{Dataset}

For collecting the dataset of WebFG 2020, we choose Bing Image Search Engine\footnote{\url{https://bing.com/}} as our web source of both training images and test images. To encourage further progress in challenging real world conditions, our dataset consists of both large numbers of categories and images. Specifically, the dataset of our challenge has 5,000 subordinate categories of different species of plants, animals and insects, as well as about 558,000 training images from web.

\begin{figure}[t]
	\centering
	\includegraphics[width=0.99\linewidth]{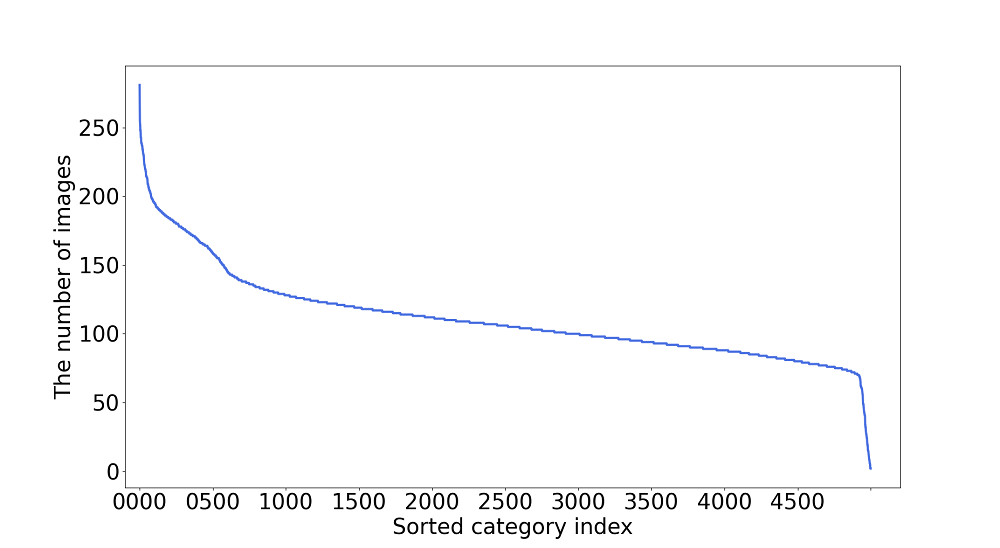}
	\caption{The data distribution of WebFG 2020.}
	\label{fig:distribution}
\end{figure}

\begin{figure*}[t]
	\centering
	\includegraphics[width=0.95\linewidth]{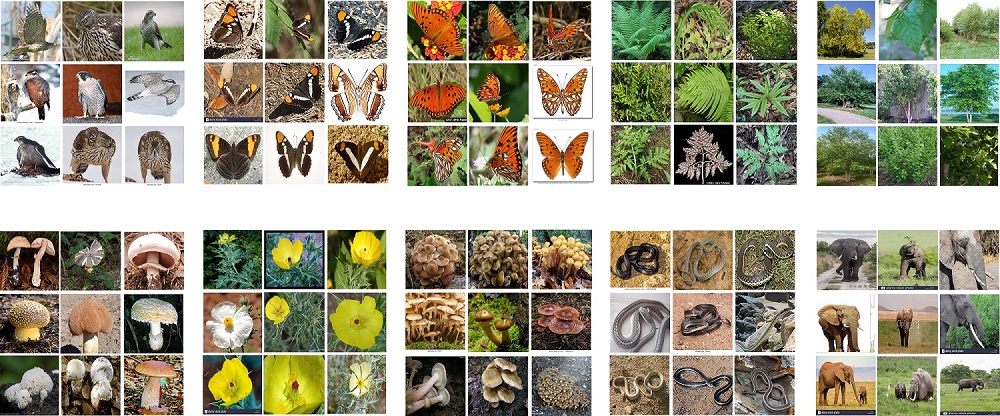}
	\caption{Exampled images of the dataset in the WebFG 2020 challenge. Each sub-figure contains nine images belonging to one sub-category. Note that, there exist data noises in these sub-categories.}
	\label{fig:examples}
\end{figure*}

Since our training images are directly crawled from the web, some broken images might be included. To remove these broken images, we employ the Python library, \ie, \texttt{Pillow},\footnote{\url{https://pillow.readthedocs.io/en/stable/}} to check each collected image and subsequently convert them to the RGB mode. Images that cannot be opened by \texttt{Pillow} or cannot be converted to RGB will be regarded as broken ones and get deleted. In consequence, we obtain 557,169 images in total for training. The data distribution of our dataset is shown in Fig.~\ref{fig:distribution}. It is apparent to observe that a long-tailed distribution phenomenon emerges in such training images. Regarding the test set, for evaluating the generalization ability, we construct a class-balanced test set (\ie, each class contains 20 images for testing) corresponding to these 5,000 sub-categories and employ the average accuracy across these 5,000 sub-categories for quantitative evaluations. All images in the test set are manually labeled, where 40\% images in the test set are used as Leaderboard-A for public evaluation and 60\% are used as Leaderboard-B for private/final evaluation. Exampled images of the dataset in the WebFG 2020 challenge are presented in Fig.~\ref{fig:examples}.

\section{Challenge}

The WebFG 2020 Challenge\footnote{\url{https://sites.google.com/view/webfg2020}} is associated with ACCV 2020\footnote{\url{http://accv2020.kyoto/}}, which is ran from Oct 9, 2020 to Nov 30, 2020. There were 54 competing teams, hailing from all over the globe. The teams came from all different backgrounds, stretching from industrial computer vision to academic ecologists, \eg, Netease, DiDi, OPPO, Huazhong University of Science and Technology, Southeast University, Tongji University, Northwestern Polytechnical University, Tianjin University, etc. More interestingly, the top ranked teams are all from industry. The best accuracy of the winner on the private test leaderboard is 71.43\%, which is significantly higher than the baseline result, \ie, 42\% on a vanilla ResNet-50.

\section{What Worked}
%\todo{In this section we can discuss themes within the winning solutions, and what was found to be the most useful. Please feel free to add any subsections here for things you found to be most helpful.}

In this section, we discuss detailed themes within the winning solutions, and what was found to be the most useful.

\subsection{Cleaning Data}

Since the challenge data are all collected from the web, severe noises exist in the data. The first thing of all winning solutions is to clean the training data, which can bring significant accuracy boosting. Particularly, the results of the third place solution report the improvement is more than 10\% on a vanilla network. 

\begin{figure*}[t]
	\centering
	\includegraphics[width=0.75\linewidth]{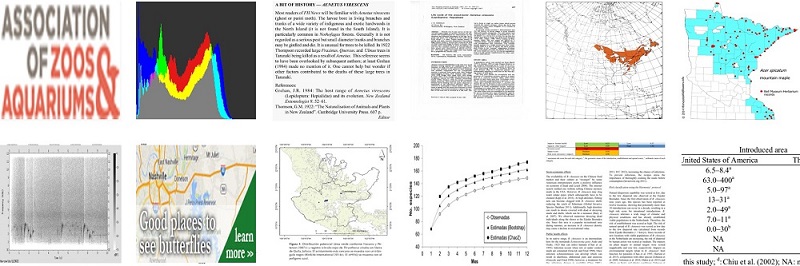}
	\caption{Irrelevant images (noisy data) in WebFG 2020.}
	\label{fig:irrelevant}
\end{figure*}

Specifically, the first place solution was to use clustering methods (cf. Sec.~\ref{sec:1st}) to generate image prototypes and remove those irrelevant images, such as images of maps, articles, charts and so on (cf. Fig.~\ref{fig:irrelevant}). The second place solution leveraged the test set to clean data since it is clean (cf. Sec.~\ref{sec:2nd}). In concretely, they treated the test set as a positive category and treated the noisy data in the training set as a negative category. A binary classification model can be trained based on the positive/negative categories. Thus, noisy data can be detected by being predicted as negative. The third place solution to clean data was based on the prediction scores of $k$-fold cross-validation (cf. Sec.~\ref{sec:3rd}). Details of these cleaning methods can be found in the supplementary materials.

\subsection{Knowledge Distillation}

It is proven that self-distillation~\cite{zhang2019your} is also effective to deal with noisy data. As aforementioned, there are a large number of inaccurate / wrong labels or ambiguity samples in the data set. In order to reduce the ambiguity of training the model, the first place solution divided all the training images into five folds, and used four of them as training and predicted the labels of the last fold repeatedly in five rounds. In each round, after model training, it obtains the predictions of the last fold, which are mixed by the ground truth labels of that fold with a certain proportion to form the modified labels. Apparently, the modified labels are the soft labels w.r.t. the corresponding noisy samples. It could relieve the severe label noises. Similarly, the third place also employed KD~\cite{gou2020knowledge}, which brought 1\% improvement.

%After model training, each fold verification set\xs{What the verification set means?} predicted to form the out-of-fold result file of all data sets, and then the out-of-fold result and the ground truth label of the data set are mixed in a certain proportion to form a new training set Labels, in other words, we assign a softened label to each image with a high probability of being a noisy sample, thereby reducing the difficulty of model training.
%\begin{figure}[t]
%	\centering
%	\includegraphics[width=1.0\linewidth]{irrelevant}
%	\begin{center}
%		\caption{Irrelevant images in our dataset.}
%	\end{center}
%	\label{fig:irrelevant}
%\end{figure}
%\subsection{Class Balanced Cleaning}

%It is also effective if removing the samples of low predicted confidence based on the modified labels. Meanwhile, it can take into account the balance of categories to reduce the probability of being mistakenly cleared for the few sample categories, so that there are not too few images in some categories and aggravate the long tail distribution. Degree, increase the difficulty of model training.

\subsection{Class Balanced Strategies}

As shown in Fig.~\ref{fig:distribution}, the distribution of the challenge data is apparently long-tailed. Thus, class balanced strategies are also crucial and effective for such a challenge. Specifically, the winning solutions employed diverse rebalancing methods to deal with the long-tailed distribution, \eg, the network architecture~\cite{bbn}, re-sampling methods~\cite{wu2020forest}, re-weighting methods~\cite{cui2019class}, etc. More details can be found in the supplementary materials.

\subsection{Bag of Tricks}

Apart from complex methods used in the challenge, simple refinements on training procedure also make contributions. These refinements, also called \emph{tricks}, are minor but effective. Among the winning solutions, some popular used tricks, \eg, mixup\cite{thulasidasan2019mixup,zhang2017mixup}, label smoothing~\cite{muller2019does}, multi-scale~\cite{gong2014multi}, various data augmentations~\cite{perez2017effectiveness}, can bring accuracy improvements. Also, some other specific tricks w.r.t. the challenge shown great potentials for accuracy boosts. Specifically, the second place solution found that it exits a prior of the challenge test data, \ie, the number of samples in each category of the test set might be the same. Therefore, for a specific class, which has more than 20 predicted samples, they will re-assign those samples whose confidence scores are less than the 20th sample’s score of that class to the second highest prediction class as their predicted class. If the second highest class already has more than 20 samples, the third highest class will be considered, and so on. The task-specific strategy is called ``class self-calibration''. This simple trick brought a pretty consistent boost of 0.7\%-1\% on both Leaderboard-A and Leaderboard-B. Meanwhile, the second place solution also observed that most images contains small-scale animals. Thus, they proposed to resize original images to a very large resolution for training to solve the small animal problem.

\section{What Did not Work}

In this section we discuss what methods were tried that did not work. Specifically, 

\begin{itemize}
\item The first place solution also tried some other denoising methods, such as AUM Ranking~\cite{pleiss2020identifying}, etc. After that, they deleted part of the images based on the outputs of AUM. However, it caused a significant accuracy drop.
\item The focal loss~\cite{lin2017focal} did not work well on the WebFG 2020 challenge data.
\item The second place solution trained a binary model that distinguishing between animals and plants. The accuracy of the binary model is very high, but there is no accuracy improvement after fusion with other models on the final recognition task.
\end{itemize}

\section{Conclusion}
In this technical report, we reviewed the overall process of the WebFG 2020 challenge, pulled together the top-3 solutions, and discussed what methods worked best across the set of winning teams, and what surprisingly did not help. Specifically, we observed that:
\begin{itemize}
\item Cleaning data and knowledge distillation are beneficial to alleviate data noises in the webly-supervised data.
\item Class balanced strategies, \eg, the tailored network architecture, re-balancing methods, can relieve the challenge of extreme class imbalance.
\item Practical tricks and some task-specific solutions (\eg, class self-calibration) also work well for WebFG 2020.
\item However, some state-of-the-art methods of denoising or imbalanced learning did not work well and even brought accuracy drops.
\end{itemize}
These observations could bring insights to the research of webly-supervised fine-grained recognition in the future. Furthermore, a series of webly-supervised fine-grained recognition challenges is encouraged to be held as a yearly competition in the fine-grained and computer vision community.

\input{acknowledgements}

\newpage
\appendix
\section*{Supplementary Materials}

In the supplementary materials, we present the detailed information of the winning solutions. The slides of these winning solutions are available at \url{https://sites.google.com/view/webfg2020/challenge}.

%\todo{I'm thinking the particular details of each solution and who was on the team could be sections in the appendix. Each team can add a file describing their solution and then we can figure out how to standardize the format.}
\input{1st_place_solution}

\input{2nd_place_solution}
\input{3rd_place_solution}

\newpage
{\small
\bibliographystyle{ieee_fullname}
\bibliography{main}
}

\end{document}

%% file: acknowledgements.tex
\section{Acknowledgements}

We sincerely thank all the participants of the WebFG 2020 challenge, and also thank Extreme Mart (belonging to Extremevision) for supporting the challenge this year. This work was supported by the Fundamental Research Funds for the Central Universities, No. 30920041111.

%% file: 1st_place_solution.tex
\section{First Place Solution}\label{sec:1st}
\subsection{Team Members (Authors of this section)}
Jia Wei, Netease Games AI Lab

Si Xi, Netease Games AI Lab

Wenyuan Xu, Netease Games AI Lab

Weidong Zhang, Netease Games AI Lab

\subsection{Method}

First of all, the important things we should do is to clean the data set. We conduct an experience on the validation set by training our model. After that, we counted 1,000 bad cases and find that there are 13\% of samples have the wrong labels. Also, there are 24\% of samples are noises. Thus, we attempt several methods to clean the data set. We use model clustering to cluster and remove the dirty samples, such as maps. Then, we use five-fold Knowledge Distillation (KD)~\cite{gou2020knowledge} to process the noise data. And then we use the new labels product by KD to remove some noisy samples, which have low confidence. We should pay attention that we have to keep class balance when we cleaned the samples with low confidences.

It is important to choose what kind of backbones we used. We use EfficientNet~\cite{tan2019efficientnet}, ResNet-based models~\cite{wu2019wider} and BBN~\cite{bbn} As the backbones. We notice that the backbone which have attention machines will get better performance. Our best single model was trained by BBN. The BBN framework is further equipped with a cumulative learning strategy~\cite{bbn}, which is designed to first learn the universal patterns and then pay attention to the tail data. BBN is a great framework to address the long tail distillation problem. But it is a kind of difficult to train, half precision, 8 V100 required one week to train. We can improve the Top-1 accuracy more than 10\%  by using BBN to train the ResNet50. We also use BBN to train another backbones.

We use a lot of tricks to train the models. We need a lot of GPUs to train BBN-based models. We use a lot of strategies to do the data augmentation and oversample. We find that focal loss was not work well in the large scale dataset. 10-crop and multi-scale were used to do the inference. mixup and label smoothing were used to improve the performance. We also use pseudo label. Specifically, we selected 50\% of the test set samples and then use KD to reduce label noises.

We take the BBN-ResNet50 as an example. The baseline accuracy is 55.7\%. After cleaning the data set, the accuracy can improve by 2.6\%. After mixup and labels smoothing is applied, the accuracy can improve by 1.5\%. After three times KD is applied, the accuracy can improve by 1.1\%. We also use many other tricks.

We have trained more than 40 models with different backbones, different input sizes and different training strategies. The best single model obtains 68.9\% accuracy. But even so, adding three weak models with 60\% accuracy can improve the final accuracy.

We have adjusted model-wise weights to boost stronger models, which can slightly improve the accuracy by 0.2\%.
All results of our 5 models are listed in the table. Finally we get 71.4\% in Leaderboard-B.
\setlength{\tabcolsep}{4pt}
\begin{table}[htbp]
	\small
	\setlength{\abovecaptionskip}{5cm}
	
	\setlength{\belowcaptionskip}{5cm}
	\begin{spacing}{1.5}
	
	\caption{Model and results in Leaderboard-B.}
	
	\end{spacing}
	
	\begin{center}
		\begin{tabular}{ccccccc}
		\hline
		Model &Accuracy(B)  \\ \hline
		BBN-SEResneXt50 &68.91 \\ \hline
		BBN-Resnet50 &66.48 \\ \hline
		Efficientnet-b3 &60.74 \\ \hline
		Efficientnet-b4 &61.26 \\ \hline
		Efficientnet-b5 &59.12 \\ \hline
		Model ensemble &71.40 \\ \hline
	\end{tabular}
\end{center}
\end{table}
\setlength{\tabcolsep}{1.4pt}

%% file: 2nd_place_solution.tex
\section{Second Place Solution}\label{sec:2nd}
\subsection{Team Members (Authors of this section)}
Xiaoxin Lv, Netease Yidun AI Lab

Dengpan Fu, Microsoft Research Asia

Qing Li, University of California, Los Angeles

Baoying Chen, Shenzhen University

Haojie Guo, Northwestern Polytechnical University

\subsection{Method}

At the beginning of the challenge, we conducted a detailed analysis of the training data and we discovered:
\begin{itemize}
\item There are a lot of noise samples in the training data set, such as maps, charts, raw texts, and so on.

\item The test data set is much cleaner than the training data set.

\item Many images of animal categories contain plants, but most pictures of plant categories do not contain animals.

\item Small animal problem: Some animals are very small in the images.
\end{itemize}

These conclusions played a great role in this challenge.

In the data processing stage, we removed about 5000 noisy data through active learning. Specifically, we used the test set as a positive category, and the noise data in the training set as a negative category. We trained an EfficientNet-B2 network and iterated three times. 

For another version of the training set, we removed about 50,000 pictures without manual verification. However, no experiments were conducted on this data set due to limitation of computing resources.

In this challenge, we choose the EfficientNet-B4/B5 with noise student pre-trained weights as our base model. Due to the huge difference between the training data and the test data, we did not divide the validation set locally and used all the data for model training. We measured model performance based on the score of Leaderboard-A. During the training process, we used cut-mix~\cite{yun2019cutmix}, auto-augment, and other data augment methods and greatly borrowed ImageNet model training hyper-parameters. We resized original images to a very large size for training to solve the small animal problem. 

Our best EfficientNet-B4 achieved a score of 63.91\% on Leaderboard-A. During the entire competition, we did not find any over-fitting phenomenon before 100 epochs, larger image size, more complex model, and more training epochs always get a better result.

Model ensemble can significantly improve the score. After merging the 4 models, we got a score of 67.275\% on Leaderboard-A. We found that the number of samples in each category of the test set is the same. For a specific class, which has more than 20 predicted samples, we will re-assign the class for those samples whose confidence scores are less than the 20th sample’s score, and we will consider the second highest class as their predicted class for those samples, if the second highest class is already has more than 20 samples, the third highest class will be considered, and so on. This simple trick gave a pretty consistent boost of 0.7\%-1\% on both Leaderboard-A and Leaderboard-B. Finally, we fine-tuned our best EfficientNet-B4 model on the old test set, after merging it with other previous 4 models, we reached 71.005\% on Leaderboard-A and 71.3\% on Leaderboard-B.

%% file: 3rd_place_solution.tex
\section{Third Place Solution}\label{sec:3rd}
\subsection{Team Members (Authors of this section)}
Zhiheng Wang, DiDi-SSTG

Taolue Xue, DiDi-SSTG

Haipeng Jing, DiDi-SSTG

Mingwen Zhang, DiDi-SSTG

Tianming Zhang, DiDi-SSTG
\subsection{Method}

As described in the first place solution and the second place solution, the most important thing is to design a method to clean the data. We take three steps to clean the data set:

First of all, training a noise data recognition model to eliminate data that is not related to the competition, such as maps, molecular structure maps, etc.

Then, according to the top5 predictions of the model, cleaning the coarse-grained label wrong data — the contents of images are obviously different, but the labels are the same.

Finally, the top1 predition of model is used to clean the fine-grained label wrong data—the contents of images is similar, but the label is the same.

After the data cleaning is completed, the accuracy of our model can be increased from 47\% to 58\%.

Next, we used a powerful data augmentation combination - mixcut, random color jitter, random direction rotation, random flip, random image quality loss, random zoom map, random grid stitching and random crop. This data augmentation combination can make our model more robust to the test set and have a stronger ability to capture detailed features. In experiments, this data augmentation combination can increase model accuracy by 3\%-4\%.

Data balance and knowledge distillation are also very important. Through simple re-weighting or re-sampling, the accuracy can be improved by 1\%. By knowledge distillation, the accuracy can also be improved by 1\%. On the other hand, a large batch size in the experiment can make the accuracy of the model converge faster and have better generalization performance on the test set. Therefore, we adopted a distributed training method and set the batch size to 2000 to train the model. Through these tricks, our single model accuracy is between 62\%-65\%. Finally, ensemble these single models, and the final recognition accuracy of Leaderboard-A is 67.818\%.